\documentclass[letter,twoside,11pt]{article}
\usepackage[latin2]{inputenc}
\usepackage[OT4]{fontenc}
\usepackage{jmlr2e}
\usepackage{matematyka_ang}
\usepackage{url}
\usepackage{graphicx}

\title{Classifying Signals With Local Classifiers}
\author{\name Wit Jakuczun \email jakuczun@mini.pw.edu.pl \\
\addr  Faculty of Mathematics and Information Science \\
Warsaw University of Technology \\
Pl. Politechniki 1 \\
00-661 Warsaw, POLAND}
\editor{}
\ShortHeadings{Classifying signals With Local Classifiers}{Wit Jakuczun}
\jmlrheading{}{}{}{}{}{Wit Jakuczun}

\begin{document}

\maketitle

\begin{abstract}%
  This paper deals with the problem of classifying signals. The new
  method for building so called \emph{local classifiers} and
  \emph{local features} is presented. The method is a combination of
  \emph{the lifting scheme} and \emph{the support vector
    machines}. Its main aim is to produce effective and yet
  comprehensible classifiers that would help in understanding
  processes hidden behind classified signals. To illustrate the method
  we present the results obtained on an artificial and  a real
  dataset.
\end{abstract}


\section{Introduction}
\label{sec:introduction}

Many classification algorithms such as artificial neural networks
induce classifiers which have good accuracy but do not give an
insight into the real process which is hidden behind the
problem. Although predictions are made with high precision such
classifiers do not answer the question ``Why?''. Even algorithms such
as decision trees or rule inducers very often produce enormous
classifiers. Their analysis is almost intractable by the human
mind. It is even worse when these algorithms are used for problems of
signal classification. In practice good accuracy without an
explanation of the classification process is useless.

In this article we describe an approach which can help in building
classifiers which are not only very accurate but also
comprehensible. The method is based on the idea of the \emph{lifting 
scheme} \citep{sweldens98lifting}. The lifting scheme
is used for calculating expansion coefficients of analysed signals
using biorthogonal wavelet bases. The biggest advantage of this method
is that it uses only spatial domain in contrast to the classical
approach \citep{daub92tenlectures} in which the frequency domain is
used. As originally lifting scheme did not give us enough freedom in
incorporating adaptation we used its modified version called
\emph{update-first} \citep{claypoole98adaptive}.

Assume we act in space $\real^N$ spanned by a biorthogonal base
$\{\phi_i\}_{i=1}^n$ and $\{\tilde{\phi}_i\}_{i=1}^n$. Vectors 
$\{\phi_i\}_{i=1}^n$ and $\{\tilde{\phi}_i\}_{i=1}^n$ are biorthogonal
in the sense that
\begin{displaymath}
  \ilskal{\phi_i}{\tilde{\phi_j}}=\delta_{ij}
\end{displaymath}
where $\delta_{ij}=1$ if $i=j$ and $0$ otherwise.

Each vector $x\in\real^N$ can be expressed in the following way
\begin{equation}
  \label{eq:biorth_expansion}
    x=\sum_{i=1}^n \alpha_i\phi_i
\end{equation}
where $\alpha_i=\ilskal{\tilde{\phi}_i}{x}$ are expansion coefficients.
Very important feature of vectors $\{\tilde{\phi}_i\}_{i=1}^n$ is
that they can be nonzero only for several indices. It implies that
for calculating $\ilskal{\tilde{\phi}_i}{x}$ only a part of the
vector $x$ is needed. This feature is called \emph{locality}.

The aim of the method presented in this article is to find such
an expansion \eqref{eq:biorth_expansion} by implicitly constructing
biorthogonal base $\{(\phi_i,\tilde{\phi}_i)\}_{i=1}^n$, that
coefficients $\alpha_i=\ilskal{\tilde{\phi}_i}{x}$ are as
discriminative as possible for classified signals. 

More specifically we assume that a training set $X=\{(x_i,y_i) :
x_i\in\real^n, y_i\in\{-1,+1\}\}_{i=1}^l$ is given. For each base
vector $\tilde{\phi}_j$ we get a vector of expansion coefficients
$\alpha^j\in\real^l$ 
\begin{displaymath}
  \alpha^j(i)=\ilskal{\tilde{\phi}_j}{x_i}
\end{displaymath}
For each such vector we can find a number $b^j\in\real$ called bias
for which
\begin{displaymath}
  \sgn(\alpha^j(i) + b^j)=y_i
\end{displaymath}
for as many indices $i\in\left\{1,2,\ldots,l\right\}$ as possible.

For calculating expansion coefficients we used the idea of
\emph{support vector machines (SVM)} introduced by
\cite{vapnik98book}\footnote{More precisely, we used PSVM a variant of SVM
  called proximal support vector machines \citep{fung01proximal}.}. SVM
proved to be one of the best classifier inducers. Combining the power
of SVM and the locality feature of the designed base we were able to
build classifiers with a very good classification accuracy and which
are also easily interpreted. We present experiments obtained for
an artificial  datasets and a real dataset. The artificial datasets
allowed us to verify our method and to better understand its features.
Experiments conducted on the real dataset proofed usefulness of the
method for real applications.

\section{Outline of the paper}
\label{sec:outline-paper}

The paper is divided into two main parts and the appendix. The first
part is devoted to a description of the method and consists of three
subparts. First we present a general outline of the method next we
introduce some notation that will be used in next part that gives
detailed description of the method. The first part of the paper we end
with a short summary of the presented method. In the second part of
the paper we present a results of the experiments conducted both on the
artificial and the real dataset. In the appendix we show how to
efficiently solve optimisation problems that arise in the method.

\section{Method description}
\label{sec:method-description}

In this section we will describe the new method for designing
discriminative biorthogonal bases for signal classification. In fact
we will be computing only expansion coefficients of some
implicitly defined discriminative biorthogonal base. The method is a
combination of \emph{update-first} version of the lifting scheme
\citep{claypoole98adaptive} and \emph{proximal support vector
  machines} \citep{fung01proximal}.

\subsection{Outline of the method}
\label{sec:emph-first-vers}

The method is based on  the Lifting Scheme that is very general and
easily modified method for computing expansion coefficients of analysed
signal with respect to biorthogonal base. The method is iterative and
each iteration is divided into three steps
\begin{itemize}
\item {\bf SPLIT} - Signal is splitted into two subsignals containing
  \emph{even} and \emph{odd} indices.
\item {\bf UPDATE} - \emph{Coarse approximation} of analysed signal is
  computed from subsignals.
\item {\bf PREDICT} - \emph{Wavelet coefficients} are calculated using
  \emph{coarse approximation} and subsignal containing \emph{even}
  indicies.  Those coefficients are simply inner products between a
  \emph{weight vector} and small part of \emph{coarse approximation} and
  \emph{even subsignal}. We used \emph{proximal support vector
    Machines} \citep{fung01proximal} to calculate the \emph{weight
    vector}. As PSVM is the procedure for generating classifiers we
  decided to call obtained expansion coefficients \emph{discriminative
    wavelet coefficients}.
\end{itemize}
\emph{Coarse approximation} is used as an input for next iteration.
As the \emph{coarse approximation} is twice shorter than original
signal the number of iterations is bounded from above by $\ln(N)$
where $N$ is the length of the analysed signal.

\subsection{Notation}
\label{sec:notation}

Assume we are given a training set $X$
\begin{displaymath}
  X=\left\{(\wektor{x}_i,y_i)\in\real^{N\times\{-1,+1\}}: i=1,\ldots,l\right\}
\end{displaymath}
where $N=2^n$ for some $n\in\nat$. Vectors $\wektor{x}_i$ are sampled
versions of signals we want to analyse and $y_i\in\{-1,+1\}$ are
labels.

Having set $X$ we create two matrices
\begin{displaymath}
  \macierz{A}=\left(
    \begin{array}{c}
      \wektor{x}_1^T \\
      \vdots\\
     \wektor{x}_l^T
    \end{array}\right)\in\real^{l\times N}
\end{displaymath}
and 
\begin{displaymath}
  \macierz{Y}=\left(
    \begin{array}{ccc}
      y_1  &       &    \\
          & \ddots &    \\
          &        & y_l
    \end{array}\right)
\end{displaymath}
Let $I=\{i_1,\ldots,i_k\}$ be a set of integer numbers (indices). We
will use the following shorthand notation for accessing indices I of a
vector $\wektor{x}\in\real^N$.
\begin{displaymath}
  \wektor{x}(I)=(\wektor{x}(i_1),\ldots,\wektor{x}(i_k))
\end{displaymath}
We will also use a special notation for accessing odd and even
indices of a vector $\wektor{x}\in\real^N$
\begin{eqnarray*}
  \wektor{x_o} &=&
  (\wektor{x}(1),\wektor{x}(3),\ldots,\wektor{x}(N-1))\quad\textrm{for
    odd indices} \\
  \wektor{x_e} &=&
  (\wektor{x}(2),\wektor{x}(4),\ldots,\wektor{x}(N))\quad\textrm{for
    even indices} \\
\end{eqnarray*}
Finally we will use the following symbols for special vectors
\begin{displaymath}
  \wektor{e}=\left(
    \begin{array}{c}
      1\\\vdots\\ 1
    \end{array}\right)
\end{displaymath}
and
\begin{displaymath}
  \wektor{e_1}=\left(
    \begin{array}{c}
      1\\0\\\vdots\\ 0
    \end{array}\right)
\end{displaymath}
The dimensionality of the vectors $\wektor{e}$ and $\wektor{e_1}$ will be
clear from the context.
\subsection{Three main steps}
\label{sec:three-main-steps}

As we have mentioned before the method we propose is iterative and
each iteration step\footnote{We also use a name
\emph{decomposition level} for iteration step.} consists of three
substeps.

\subsubsection{\bf First substep - Split}
\label{sec:split}

Matrix $\macierz{A}$ is splitted into matrices $\macierz{A_o}$ (odd
columns) and $\macierz{A_e}$ (even columns)
\begin{displaymath}
  \macierz{A_o}=\left(
    \begin{array}{c}
      \wektor{x_o}_1^T \\
      \vdots\\
      \wektor{x_o}_l^T
    \end{array}\right)\in\real^{l\times N/2}
\end{displaymath}
and
\begin{displaymath}
  \macierz{A_e}=\left(
    \begin{array}{c}
      \wektor{x_e}_1^T \\
      \vdots\\
      \wektor{x_e}_l^T
    \end{array}\right)\in\real^{l\times N/2}
\end{displaymath}

\subsubsection{\bf Second substep - Update}
\label{sec:update}

Having matrices $\macierz{A_o}\in\real^{l\times N/2}$ and
$\macierz{A_e}\in\real^{l\times N/2}$ we create matrix
$\macierz{C}\in\real^{l\times N/2}$
\begin{displaymath}
  \macierz{C}=\frac{1}{2}\left(\macierz{A_o} + \macierz{A_e}\right)=\left(
  \begin{array}{c}
    \wektor{c}_1^T \\
    \vdots \\
    \wektor{c}_l^T
  \end{array}\right)
\end{displaymath}
This matrix will be called \emph{coarse approximation} of matrix $\macierz{A}$.

\subsubsection{\bf Third substep - Predict}
\label{sec:predict}

In the last step we calculate \emph{discriminative wavelet
coefficients}. For each column $k$ of matrix $\macierz{A_e}$
($k=1,2,\ldots,N/2$) we create matrix $\macierz{A}^k\in\real^{l\times
  L_k+1}$ where $L_k\in\nat$ is an even number and a parameter of
the method.\footnote{In presented experiments we assumed that $L_k=L$
  for some constant $L\in\nat$.} 
\begin{displaymath}
  \macierz{A}^k=\left(
    \begin{array}{cc}
      \wektor{x_e}_1(k) & -\wektor{c}^k_1 \\
      \vdots & \vdots \\
      \wektor{x_e}_l(k) & -\wektor{c}^k_l
    \end{array}\right)
\end{displaymath}
where $\wektor{c}^k_i=\wektor{c}_i(I_k)$ and $I_k$ is a set of indices
selected in the following way 
\begin{itemize}
\item If $1\leq k < \frac{L_k}{2}$ then $I_k=\{1,2,\ldots,L_k\}$
\item If $\frac{L_k}{2} \leq k < \frac{N}{2}-\frac{L_k}{2}$ then
  $I_k=\{k-\frac{L_k}{2}+1,\ldots,k+\frac{L_k}{2}\}$ 
\item If $\frac{N}{2} - \frac{L_k}{2} \leq k \leq \frac{N}{2}$ then
  $I_k=\{\frac{N}{2}-L_k+1,\ldots,\frac{N}{2}\}$ 
\end{itemize}

At this point our method can be splitted into two variants:
regularised and non-regularised.

\begin{itemize}
\item {\bf regularised variant:} This variant uses PSVM
  approach to find the optimal \emph{weight vector}
  $\wektor{w}^k\in\real^{L_k+1}$. According to \cite{fung01proximal}
  optimal $\wektor{w}^k$  is the solution of the following
  optimisation problem
  \begin{equation}
    \label{eq:reg-opt-1}
    min_{\wektor{w}^k,\gamma_k,\wektor{\xi}^k}
    \frac{1}{2}\norm{\wektor{w}^k}^2_2 + \frac{1}{2}\gamma_k^2 +
    \frac{\nu_k}{2}\norm{\wektor{\xi}_k}^2_2
  \end{equation}
  subject to constraints
  \begin{equation}
    \label{eq:reg-opt-2}
    \macierz{Y}(\macierz{A}^k\wektor{w}^k -
    \gamma_k\wektor{e})+\wektor{\xi}^k=\wektor{e}
  \end{equation}
  where $\xi^k$ is the error vector and $\nu^k\geq 0$. 
\item {\bf non-regularised variant:} Similarly as in \emph{regularised}
  variant the optimal \emph{weight vector}
  $\wektor{w}^k\in\real^{L_k}$ is given by solving the following
  optimisation problem
  \begin{equation}
    \label{eq:nreg-opt-1}    
    min_{\wektor{w}^k,\gamma_k,\wektor{\xi}^k}
    \frac{1}{2}\norm{\wektor{w}^k}^2_2 + \frac{1}{2}\gamma_k^2 +
    \frac{\nu_k}{2}\norm{\wektor{\xi}_k}^2_2
  \end{equation}
  subject to constraints
  \begin{equation}
    \label{eq:nreg-opt-2}
    \macierz{Y}\left(\macierz{A}^k\left(1\atop \wektor{w}^k\right) -
    \gamma_k\wektor{e}\right)+\wektor{\xi}^k=\wektor{e}
  \end{equation}
  where $\xi^k$ is the error vector and $\nu^k\geq 0$. The only difference
  to the previous variant is that dimensionality of $\wektor{w}^k$ is
  $L_k$ instead of $L_k+1$ and $x_{ei}(k)$ is multiplied by one.

  In this variant we can also add some extra constraints such
  that in case of polynomial signals (up to some degree $p_k$) we will
  get wavelet coefficients equal to zero. These constraints can be
  written in the following way
  \begin{equation}
    \label{eq:nreg-opt-3}
    \macierz{B}^k\wektor{w}^k=\wektor{e_1}
  \end{equation}
  where $\wektor{e_1}\in\real^{p_k}$ and $\macierz{B}^k$ consists of
  the first $p_k$ rows of the Vandermonde matrix for some knots
  $t_1,t_2,\ldots,t_{L_k}$. For more details on how to select knots we
  refer reader to \citep{claypoole98adaptive} and \citep{fps:spie96}.
 
  The additional constraints could be useful if analysed signals are
  superposition of polynomial and some other possibly
  \emph{interesting component}.  They imply that polynomial part
  of the analysed signal is eliminated and thus \emph{interesting
    component} will play a bigger role in constructing
  \emph{discriminative wavelets coefficients}. Also constructed base
  will have similar properties to the standard wavelet base. In the
  appendix the reader can find information on how to efficiently solve
  this extended optimisation problem. We have not used this variant in
  our experiments but present it for completeness reasons. 
\end{itemize}

Having optimal weight vector $\wektor{w}^k$ we can calculate vector
$\wektor{d}^k\in\real^l$ of \emph{discriminative wavelet coefficients}
using the following equations
\begin{itemize}
\item {\bf regularised variant}
  \begin{displaymath}
    \wektor{d}^k(i)=\ilskal{\wektor{w}^k}{\left(\wektor{x_e}_i(k)
        \atop -\wektor{c}^k_i\right)}\quad i=1,2,\ldots,l
  \end{displaymath}
\item {\bf non-regularised variant}
  \begin{displaymath}
    \wektor{d}^k(i)=\wektor{x_e}_i(k) -
    \ilskal{\wektor{w}^k}{\wektor{c}^k_i}\quad i=1,2,\ldots,l
  \end{displaymath}  
  where $\ilskal{\cdot}{\cdot}$ is a standard inner product.
\end{itemize}
In a result we obtain a matrix $\macierz{D}\in\real^{l\times N/2}$
\begin{displaymath}
  \macierz{D}=\left(
  \begin{array}{ccc}
    \wektor{d}^1 & \cdots & \wektor{d}^{N/2}
  \end{array}\right)
\end{displaymath}

\subsection{Iteration step}
\label{sec:iteration-step}

The whole algorithm can be written in the following form
\begin{itemize}
\item Let $M$ be the number of iterations (decomposition
  levels).
\item Let $\macierz{A}_0$=$\macierz{A}$ 
\item For $m=1,\ldots,M$ do
  \begin{itemize}
  \item Calculate $\macierz{C}_m\in\real^{l\times\frac{N}{2^m}}$ and
      $\macierz{D}_m\in\real^{l\times\frac{N}{2^m}}$ by applying three
        steps described in the previous section to the matrix
        $\macierz{A}_{m-1}$.
  \item Set $\macierz{A}_m=\macierz{C}_m$. 
  \end{itemize}
\end{itemize}

The output of the algorithm will be a set of matrices
$\macierz{C}_M,\macierz{D}_1,\ldots,\macierz{D}_M$. On the basis of
these matrices we create the new training set
\begin{equation}
  \label{eq:X_new}
  X^{new}=\left\{(\wektor{x}^{new}_i,y_i)\in\real^{N\times\{-1,+1\}} :  i=1,\ldots,l\right\}
\end{equation}
where new examples are created by merging rows of matrices
$\macierz{C}_M,\macierz{D}_1,\ldots,\macierz{D}_M$. 

\subsection{Method summary}
\label{sec:short-summary}

We introduced the method that maps the set of signals $X$ into a new
set of signals $X^{new}$. In the presented setting this map is a
linear and invertible function $f\colon\real^N\to\real^N$
\begin{displaymath}
  f(x)=(\wektor{c}_M^T,\wektor{d}_1^T,\ldots,\wektor{d}_M^T)
\end{displaymath}
where
\begin{eqnarray*}
  \wektor{c}_M&\in&\real^\frac{N}{2^M}\\
  \wektor{d}_M&\in&\real^\frac{N}{2^M}\\
  &\vdots&\\
  \wektor{d}_2&\in&\real^\frac{N}{4}\\
  \wektor{d}_1&\in&\real^\frac{N}{2}\\
\end{eqnarray*}
are calculated by the method. With increasing $m$ more and
more samples from the original signal is used to calculate expansion
coefficients. For example if we set $L_k\equiv L$ for all
$k$ then to calculate vector $\wektor{d}^k$ $L2^m$ samples of the
original signal will be used.

Here we present two most important features of the method
\begin{itemize}
\item Motivation for the method is that only a small part of the
  signals is important in classification process. The method tries to
  identify this important part adaptively.
\item Exploiting natural parallelism (calculating $\wektor{d}^k$ is
  completely independent for each $k$) and Sherman-Morrison-Woodbury
  formula \citep{golubloan96:matrix_computations} the method can be
  implemented very efficiently. In the appendix
  \ref{sec:effic-solv-optim} we show how to properly solve
  optimisation problems that appears in our method.
\end{itemize}


\section{Applications}
\label{sec:applications}

This section contains description of possible applications of the
proposed method. It is divided into two parts. In the first part we
present an illustrative example of analysing artificial signals with
the proposed method. In the second part we present the results for the real
dataset.

\subsection{Artificial datasets}

Here we present results obtained on artificial datasets:
\emph{Waveform} and \emph{Shape}.

\label{sec:artif-datas-illustr}
\subsubsection{Dataset description}
\label{sec:dataset-description}

Waveform is a three class artificial dataset
\citep{breiman98arcing}. For our experiments we used a slightly
modified version \citep{saito94phd}. Three classes of signals were
generated using the following formulas 
\begin{eqnarray}
  \label{eq:wave_class_1}
  x^1(i)&=&uh_1(i)+(1-u)h_2(i)+\epsilon(i)\quad\textrm{class 1} \\
  \label{eq:wave_class_2}
  x^2(i)&=&uh_1(i)+(1-u)h_3(i)+\epsilon(i)\quad\textrm{class 2} \\
  \label{eq:wave_class_3}
  x^3(i)&=&uh_2(i)+(1-u)h_3(i)+\epsilon(i)\quad\textrm{class 3} \\
\end{eqnarray}
where $i=1,2,\ldots,32$, $u$ is a uniform random variable on the
interval $(0,1)$, $\epsilon(i)$ is a standard normal variable and
\begin{eqnarray*}
  h_1(i) &=& max(6-|i-7|,0) \\
  h_2(i) &=& h_1(i-8) \\
  h_3(i) &=& h_1(i-4)
\end{eqnarray*}

\subsubsection{Analysis}
\label{sec:waveform_analysis}

\begin{figure}[!hbtp]
  \centering
  \includegraphics[angle=270, scale=0.4]{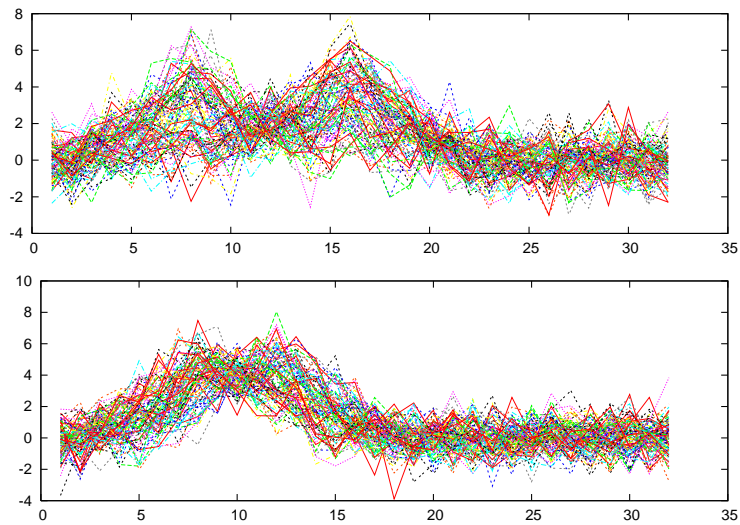}
  \caption{Examples from classes 1 and 2 }
  \label{fig:waveform_examples}
\end{figure}

For simplicity reasons we decided to concentrate only on classes
1 and 2 presented in the Figure \ref{fig:waveform_examples}. For the
purpose of this presentation we set parameters of our method as
follows
\begin{eqnarray*}
  L_k   &=& 4\\
  \nu_k &=& 1\\
  M     &=& 3
\end{eqnarray*}

\begin{figure}[!hbtp]
  \centering
  \includegraphics[angle=270, scale=0.5]{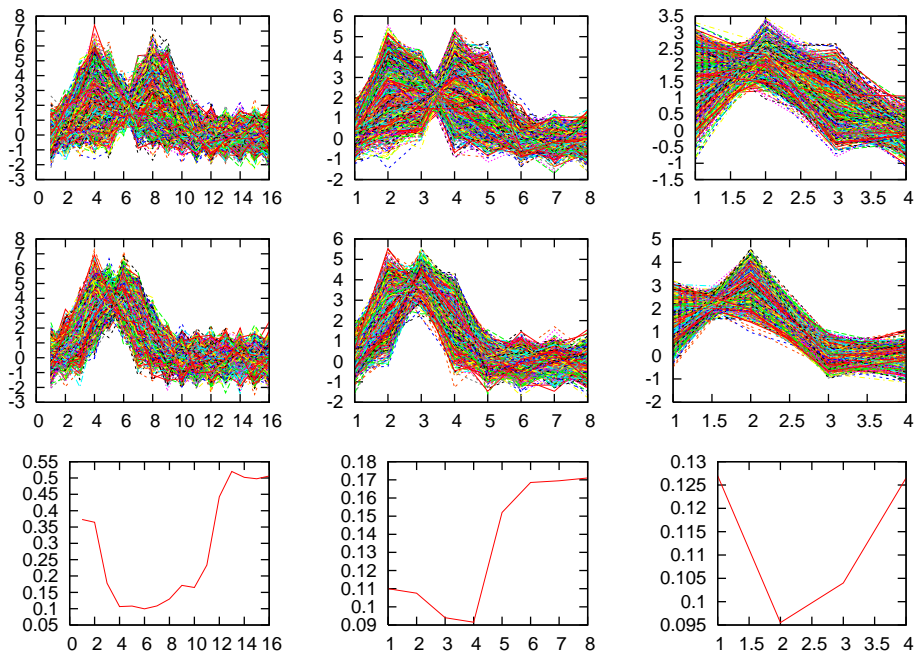}
  \caption{Coarse approximations (two upper rows) and test error of
    \emph{discriminative wavelet coefficients} (third row) for
    examples from classes 1 and 2.} 
  \label{fig:waveform_decomp}
\end{figure}

\begin{figure}[!hbtp]
  \centering
  \includegraphics[angle=270, scale=0.5]{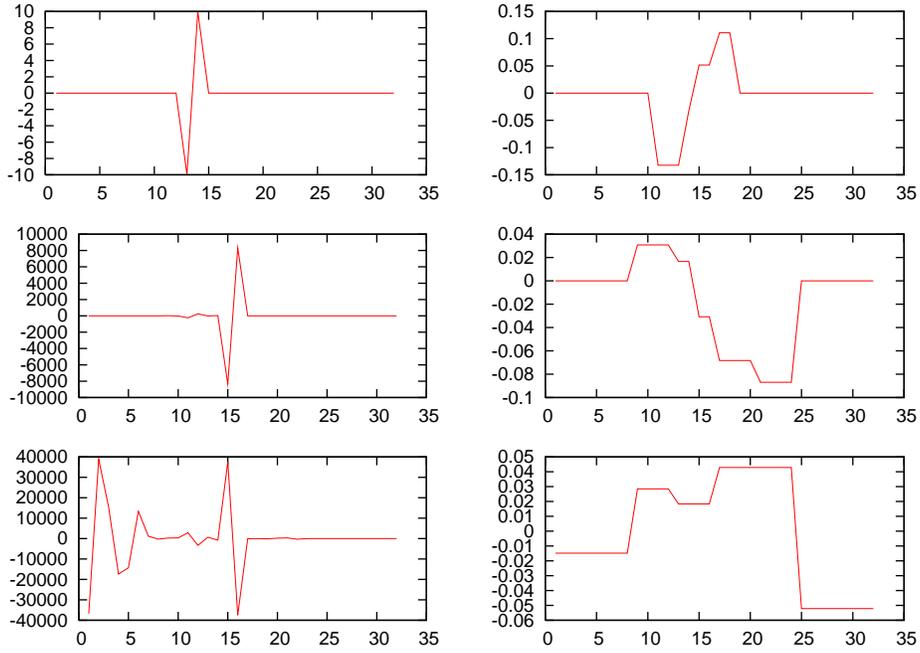}
  \caption{Best synthesis (left) and analysis (right) base vectors for
    each decomposition level}
  \label{fig:waveform_bestbase}
\end{figure}

Figure \ref{fig:waveform_decomp} presents coarse approximations (the
first two rows) and the test error ratio (the third row)
\footnote{Test error ratio obtained using all samples was equal
  $0.10$.} of calculated \emph{discriminative wavelet coefficients}
(evaluated on a separate test set). Each column present distinct
\emph{decomposition level} of 
our method. It is easily seen that \emph{coarse approximations} are
an averaged and a shortened versions of original signals. We believe
that in some cases such averaging could be very useful especially when
the analysed signals contains much noise. From the last row of the Figure
\ref{fig:waveform_decomp} we can deduce that the classification ratio of
some \emph{discriminative wavelet coefficients} is comparable to
the classification ratio obtained by applying PSVM method to the original
dataset. We can point out explicitly the period of time in which two
classes of signals differ most. This feature we called
\emph{locality}. Let us take a closer look at the $6$th
\emph{discriminative wavelet coefficient} from the first
\emph{decomposition level}. To calculate this coefficient we need $8$
out of $32$ samples of analysed signals (see first row of the Figure
\ref{fig:waveform_bestbase}). 

In the Figure \ref{fig:waveform_bestbase} one can see that base analysis
vectors with the lowest error ratio have the supports shorter than their
length. This means that to discern classes 1 and 2 we do not need all
32 samples but only a small fraction of them. Moreover when comparing
Figures \ref{fig:waveform_examples} and \ref{fig:waveform_bestbase} it
is clear that best analysis base vectors are nonzero where supports of
functions $h_1$ and $h_3$ intersect and this is the place where
analysed signals indeed differ.

\begin{figure}[!hbtp]
  \centering
  \includegraphics[width=12cm, height=8cm]{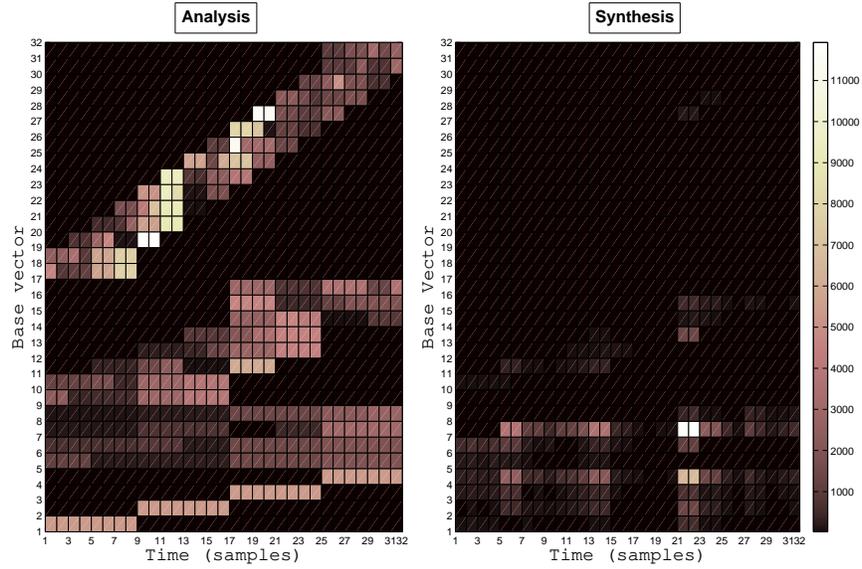}
  \caption{Supports of analysis and synthesis discriminative base}
  \label{fig:waveform_supports}
\end{figure}

The last Figure \ref{fig:waveform_supports} shows supports of analysis
and synthesis base vectors. It is easily seen that support of a base
vector widens with decomposition level. 

\subsubsection{Extracting new features}
\label{sec:waveform-extr-new-feat}

The method we presented can also be used as a \emph{supervised feature
  extractor}. Instead of feeding classifier with original training
set $X$ we use $X^{new}$ defined in \eqref{eq:X_new}. Table
\ref{tab:waveform-feature-extract} contains results of replacing
original data with new features for classifying \emph{Waveform}
dataset and \emph{Shape} dataset \citep{saito94phd} with C4.5
classifier \citep{weka99book}. From this table we can derive that
classification ratio increased considerably. We have also noticed a
substantial decrease of decision tree complexity. As our method is
designed for two-class problems and the used datasets are three-class
problems we used \emph{one-against-one} scheme
\citep{oai:CiteSeerPSU:537288}.
\begin{table}[!htbp]
  \centering
  \begin{tabular}{|r||c|c|}
      \hline
      {\bf Dataset} & \multicolumn{2}{|c|}{\bf Misclassification ratio} \\
      & Original & New  \\\hline\hline
      Waveform & $0.290$ & $0.186$ \\\hline
      Shape & $0.081$ & $0.023$ \\\hline
  \end{tabular}
  \caption{Effect of feature extraction for C4.5. Numbers are
    misclassification ratios.}
  \label{tab:waveform-feature-extract}
\end{table}

\subsubsection{Ensemble of local classifiers }
\label{sec:waveform-ensemble-local-class}

The coefficients calculated by our method can also be used directly for
classification. Table \ref{tab:waveform-voting} contains the test error
ratios for \emph{Waveform} and \emph{Shape} datasets obtained by
voting few best coefficients. As in the previous experiment we used
\emph{one-against-one} scheme for decomposing multi-class problems
into three two-class problems.

\begin{table}[!htbp]
  \centering
  \begin{tabular}{|r||c|c|c|}
      \hline
      {\bf Dataset} & \multicolumn{3}{c|}{\bf Misclassification ratio} \\
      & $3$ coefficients & $15$ coefficients & PSVM\\
      \hline\hline
      Waveform & $0.155$& $0.147$ & $0.193$\\\hline
      Shape & $0.034$ & $0.032$ & $0.094$\\\hline
  \end{tabular}
  \caption{Misclassification ratios for voting scheme. We were
    combining 3 and 15 coefficients. The last column shows
    the misclassification ratio obtained using PSVM and all samples.}
  \label{tab:waveform-voting}
\end{table}

\subsubsection{Conclusions }
\label{sec:waveformconclusions}

The presented method give both accurate and comprehensible solution
to classification problems. It can be very useful not only as a
classifier inducer but also as source of information about classified
signals. In the next section we support our claims with presenting the
results obtained on the real dataset.

\subsection{Classifying evoked potentials}
\label{sec:class-evok-potent}

In this section we present the results obtained on the dataset
collected in Nencki Institute of Experimental Biology of Polish
Academy of Science. 
The dataset consists of sampled evoked potentials of rat's brain
recorded in two different conditions. As a result the dataset consists
of two groups of recordings (CONTROL and COND) that represent two
different states of the rat's brain.  
The aim of the experiment was to explain the differences between
the two groups. We refer the reader to \cite{kublik01:pca} and
\cite{wypych03:sortingsbywavelets} for more details and previous
approaches to the data.

It should be mentioned that the problem is not a typical
classification task. This is due to the following reasons
\begin{itemize}
\item Each example (evoked potential) is labelled with an unknown noise. It
  means that there are examples that are possibly incorrectly labelled.
\item The problem is ill-conditioned due to a small number of examples
  (45-100) and a huge dimension (1500 samples).
\item The biologists that collected the data were interested not only
  in a good classification ratio but also in explanation of differences
  in the two groups.
\end{itemize}

\begin{figure}[!hbtp]
  \centering
  \includegraphics[width=15cm, height=8cm]{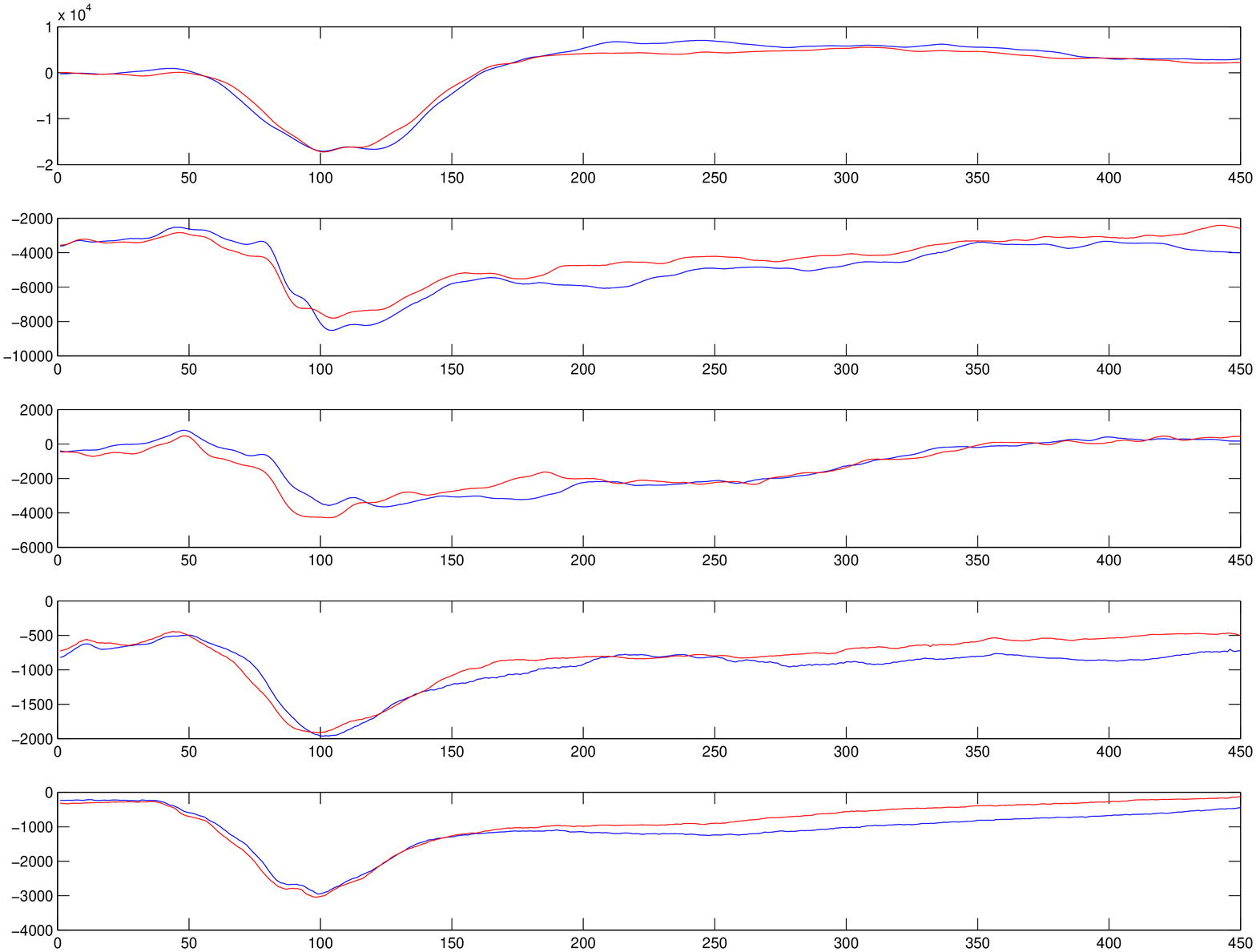}
  \caption{Averaged evoked potentials for five rats. Red colour
    denotes COND and blue denotes CONTROL. Only first most
    informative 45ms (450 samples) are presented.}
  \label{fig:ep_all_averages}
\end{figure}

Figure \ref{fig:ep_all_averages} presents averaged potentials from two
classes for group of five rats. We show only the first 45ms because
differences in this period of time can be easily interpreted by
biologists.

\begin{figure}[!hbtp]
  \centering
  \includegraphics[width=15cm, height=8cm]{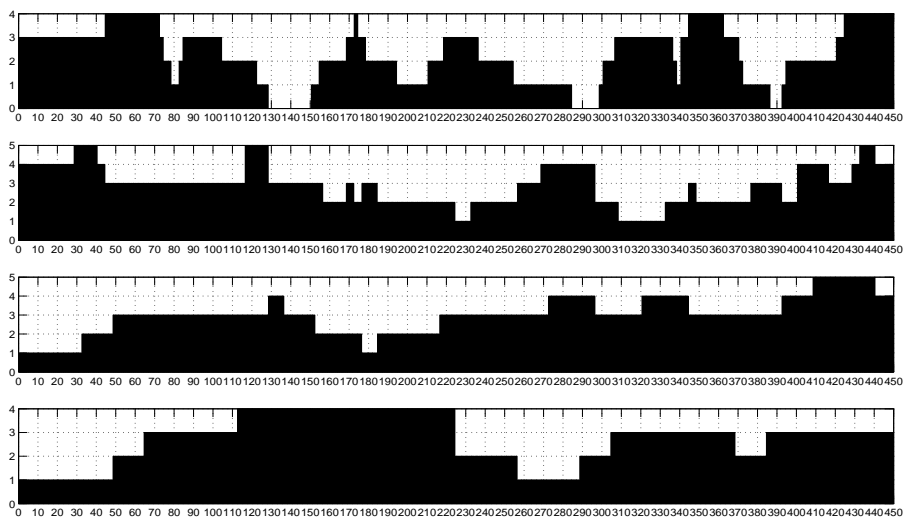}
  \caption{Histograms showing which parts of analysed signals are
    commonly indicated for all rats. The picture shows first four
    levels of decomposition of our method.}
  \label{fig:ep_all_supports}
\end{figure}

\begin{figure}[!hbtp]
  \centering
  \includegraphics[width=15cm, height=8cm]{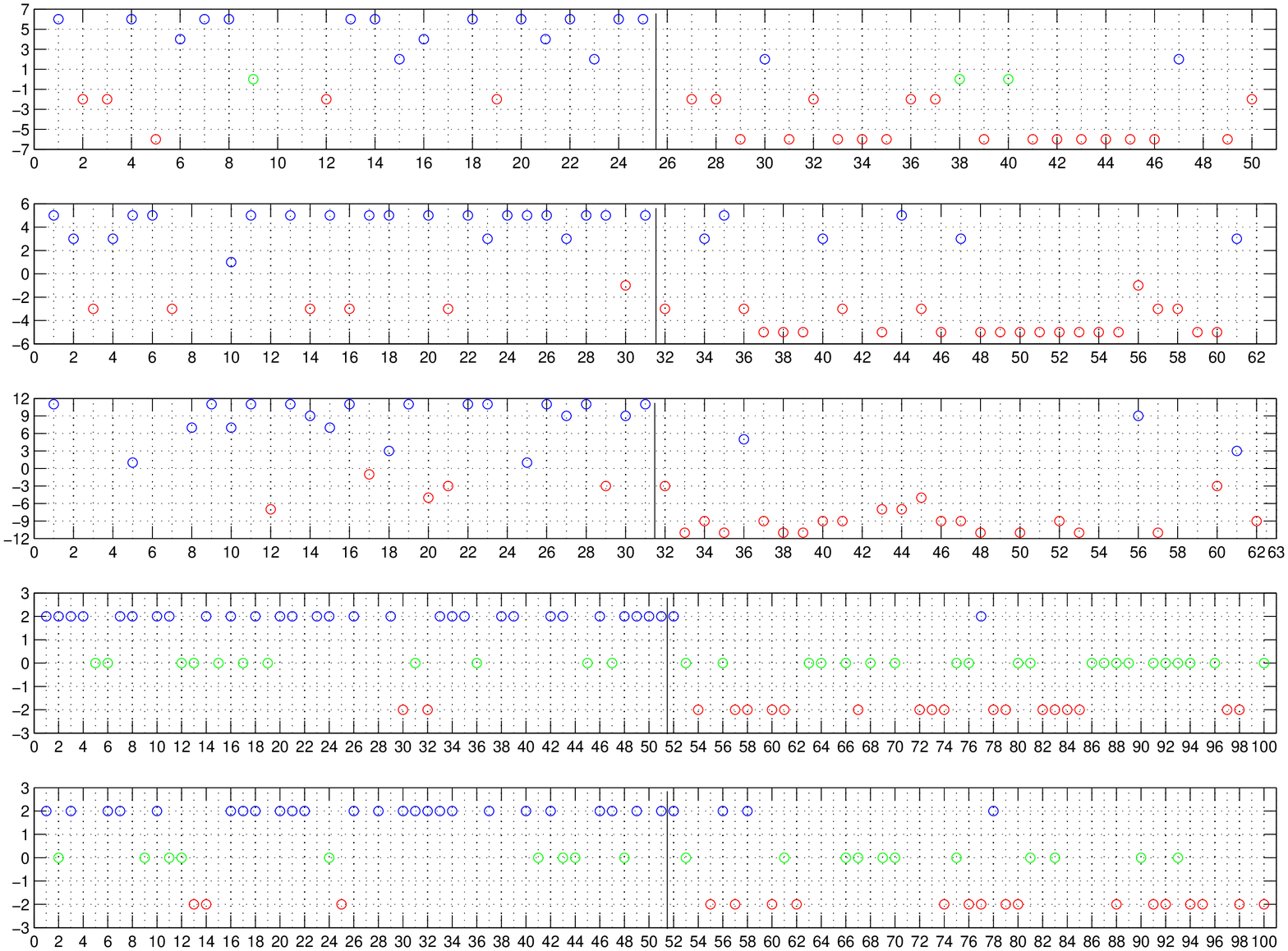}
  \caption{Charts presenting how particular potential was
    classified by selected local classifiers. Vertical line 
    divides potentials into two groups (CONTROL is on the left, COND
    is on the right). } 
  \label{fig:ep_all_sortings}
\end{figure}

After applying our method to evoked potentials for each rat we have
chosen those local classifiers whose classification accuracy was
greater or equal $0.75$ and it was statistically significant at the
level $0.1$ with respect to permutation tests
\citep{wypych03:sortingsbywavelets}. The result of this selection is
depicted in the Figure \ref{fig:ep_all_supports}. It is clear that the
most interesting parts of the signals are $2.9$-$4$ms and
$11.7$-$12.8$ms. Figure \ref{fig:ep_all_sortings} shows
how each potential is classified by selected local classifiers. It
should be read in the following manner
\begin{itemize}
\item Vertical line divides potentials into two groups CONTROL (on
  the left) and COND (on the right).
\item Axis Y shows how selected classifiers agreed on classifying
  potential.
\item The potentials were grouped (red and blue) depending on how they
  were classified. Those marked with green colour could not be
  classified.
\item We claim that those groups shows two different states of the
  rat's brain.
\end{itemize}

The presented method gave very similar results to the previous
approaches \citep{kublik01:pca}, \citep{wypych03:sortingsbywavelets}
and \citep{smolinski:rsctc2002}. Thanks to \emph{locality} feature of
our method we were able not only to classify potentials but also to
point out the most informative part of the signals. For
detailed physiological interpretation of the results we refer
the reader to \cite{wit05:class_evoked_poten}.

\section{Conclusions}
\label{sec:conclusions}

In this article we presented a new method for classifying signals. 
The method is iterative and adapts to local structures of analysed
signals. If carefully implemented it can be very efficient and when
used by an experienced researcher can be a very powerful tool for
signals discriminative analysis. There are many possible extensions to
our method but the most interesting seem to be the following
\begin{itemize}
\item Modification of the method to handle two dimensional signals such us
  images.
\item Applying \emph{kernel trick} in constructing local
  classifiers. That would lead to nonlinear classifiers and possibly
  better accuracy.
\item Constructing classifiers using \emph{Multi Kernel Learning}
  approaches \citep{bachatal:mkl}.
\end{itemize}

\appendix

\section{Efficiently solving optimisation problem for non-regularised
  and regularised version}
\label{sec:effic-solv-optim}

Here we explain how to efficiently solve optimisation problem
defined by \eqref{eq:nreg-opt-1}, \eqref{eq:nreg-opt-2},
\eqref{eq:nreg-opt-3}. Let us write Lagrangian for the optimisation
problem
\begin{eqnarray*}
  L(\wektor{w}^k,\gamma_k,\wektor{\xi}^k,\wektor{u}^k,\wektor{v}^k)&=&
  \frac{1}{2}(\norm{\wektor{w}^k}^2_2+ \gamma_k^2) +
  \frac{\nu_k}{2}\norm{\wektor{\xi}^k}^2_2 +\\
  &-& (\wektor{u}^k)^T\left(\macierz{Y}\left(\macierz{A}^k\left(1\atop
        \wektor{w}^k\right) -
      \gamma_k\wektor{e}\right)+\wektor{\xi}^k-\wektor{e}\right) \\
  &-& (\wektor{v}^k)^T\left(\macierz{B}^k\wektor{w}^k-\wektor{e_1}\right)
\end{eqnarray*}
where $\wektor{u}^k\in\real^l$ is the Lagrange multiplier associate
with the equality constraint \eqref{eq:nreg-opt-2} and
$\wektor{v}^k\in\real^{p_k}$ is the Lagrange multiplier associated
with the equality constraint \eqref{eq:nreg-opt-3}.

Settings the gradients of L to zero we get the following optimality
conditions
\begin{eqnarray}
  \label{eq:nreg_opt_app_1}
  \wektor{w}^k &=& (\macierz{\tilde{\tilde{A}}}^k)^T\macierz{Y}\wektor{u}^k -
  (\macierz{B}^k)^T\wektor{v}^k\\
  \label{eq:nreg_opt_app_2}
  \gamma_k &=& -\wektor{e}^T\macierz{Y}\wektor{u}^k \\
  \label{eq:nreg_opt_app_3}
  \wektor{\xi}^k &=& \frac{1}{\nu^k}\wektor{u}^k \\
  \label{eq:nreg_opt_app_4}
  \macierz{Y}\left(\macierz{\tilde{A}}^k +\macierz{\tilde{\tilde{A}}}^k\wektor{w}^k -
    \gamma_k\wektor{e}\right)+\wektor{\xi}^k &=& \wektor{e} \\
  \label{eq:nreg_opt_app_5}
  \macierz{B}^k\wektor{w}^k&=&\wektor{e_1}
\end{eqnarray}
where $\macierz{A}^k=\left(\macierz{\tilde{A}}^k \macierz{\tilde{\tilde{A}}}^k\right)$

Substituting \eqref{eq:nreg_opt_app_1} into \eqref{eq:nreg_opt_app_5}
we get
\begin{equation}
  \label{eq:v_k}
  \wektor{v}^k=\left[\macierz{B}^k\left(\macierz{B}^k\right)^T\right]^{-1}\left(\macierz{B}^k\left(\macierz{\tilde{\tilde{A}}}^k\right)^T\macierz{Y}\wektor{u}^k - \wektor{e}\right)
\end{equation}
Substituting \eqref{eq:nreg_opt_app_1}, \eqref{eq:nreg_opt_app_2},
\eqref{eq:nreg_opt_app_3} and \eqref{eq:v_k} into
\eqref{eq:nreg_opt_app_4} we get
\begin{equation}
  \label{eq:u_k_1}
  \macierz{Y}\left\{\macierz{\tilde{\tilde{A}}}^k\left(\macierz{\tilde{\tilde{A}}}^k\right)^T\macierz{Y}\wektor{u}^k - \macierz{\tilde{\tilde{A}}}^k\left(\macierz{B}^k\right)^T\left[\macierz{B}^k\left(\macierz{B}^k\right)^T\right]^{-1}\left(\macierz{B}^k\left(\macierz{\tilde{\tilde{A}}}^k\right)^T\macierz{Y}\wektor{u}^k - \wektor{e}\right)\right\} + \frac{1}{\nu^k}\wektor{u}^k=\wektor{e} - \macierz{Y}\macierz{\tilde{A}}^k
\end{equation}
Simplifying \eqref{eq:u_k_1} we get
\begin{equation}
  \label{eq:u_k_2}
  \begin{array}{rcl}
    \macierz{Y}\left\{\macierz{\tilde{\tilde{A}}}^k\left(\macierz{\tilde{\tilde{A}}}^k\right)^T - \macierz{\tilde{\tilde{A}}}^k\left(\macierz{B}^k\right)^T\left[\macierz{B}^k\left(\macierz{B}^k\right)^T\right]^{-1}\macierz{B}^k\left(\macierz{\tilde{\tilde{A}}}^k\right)^T\right\}\macierz{Y}\wektor{u}^k + \frac{1}{\nu^k}\wektor{u}^k &=& \\
    = \wektor{e} -
    \macierz{Y}\macierz{\tilde{A}}^k-\macierz{\tilde{\tilde{A}}}^k\left(\macierz{B}^k\right)^T\left[\macierz{B}^k\left(\macierz{B}^k\right)^T\right]^{-1}\wektor{e} &&
  \end{array}  
\end{equation}
Let matrix $\macierz{H}_1^k$ be defined as
\begin{equation}
  \label{eq:matrix_H_1}
  \macierz{H}_1^k=\macierz{Y}\left[\macierz{\tilde{\tilde{A}}}^k | -
    \macierz{\tilde{\tilde{A}}}^k\left(\macierz{B}^k\right)^T\left(\macierz{C}^k\right)^T\right]
\end{equation}
and matrix $\macierz{H}_2^k$ be defined as
\begin{equation}
  \label{eq:matrix_H_2}
  \macierz{H}_2^k=\macierz{Y}\left[\macierz{\tilde{\tilde{A}}}^k |
    \macierz{\tilde{\tilde{A}}}^k\left(\macierz{B}^k\right)^T\left(\macierz{C}^k\right)^T\right]  
\end{equation}
where
\begin{displaymath}
  \left[\macierz{B}^k\left(\macierz{B}^k\right)^T\right]^{-1}=\left(\macierz{C}^k\right)^T\macierz{C}^k
\end{displaymath}
Rewriting equation \eqref{eq:u_k_2} we obtain that
\begin{equation}
  \label{eq:u_k_3}
  \left(\frac{1}{\nu_k}\macierz{I} +
    \macierz{H}_1\left(\macierz{H}_2\right)^T\right)\wektor{u}^k=\wektor{e} -
\macierz{Y}\macierz{\tilde{A}}^k-\macierz{\tilde{\tilde{A}}}^k\left(\macierz{B}^k\right)^T\left[\macierz{B}^k\left(\macierz{B}^k\right)^T\right]^{-1}\wektor{e}
\end{equation}
Setting vector $\wektor{b}^k=\wektor{e} -
\macierz{Y}\macierz{\tilde{A}}^k-\macierz{\tilde{\tilde{A}}}^k\left(\macierz{B}^k\right)^T\left[\macierz{B}^k\left(\macierz{B}^k\right)^T\right]^{-1}\wektor{e}$
we get that vector $\wektor{u}^k$ is given by the following set of equations
\begin{equation}
  \label{eq:u_k_4}
  \left(\frac{1}{\nu_k}\macierz{I} +
    \macierz{H}_1\left(\macierz{H}_2\right)^T\right)\wektor{u}^k=\wektor{b}^k
\end{equation}
Solving above set of equations is very expensive as the number of
equations is equal to number of training examples $l$ which can be
large. Using the Sherman-Morrison-Woodbury formula
\citep{golubloan96:matrix_computations} we can calculate
$\wektor{u}^k$ as follows
\begin{equation}
  \label{eq:u_k_5}
  \wektor{u}^k=\nu_k\left(\macierz{I} -
    \macierz{H}_1\left(\frac{1}{\nu_k}\macierz{I}+\left(\macierz{H}_1\right)^T\macierz{H}_2\right)^{-1}\left(\macierz{H}_2\right)^T\right)\wektor{b}^k
\end{equation}
It should be stressed that using equation \eqref{eq:u_k_5} for computing
$\wektor{u}_k$ is much less expensive than using equation
\eqref{eq:u_k_4} because the dimensions of matrix 
\begin{displaymath}
  \frac{1}{\nu_k}\macierz{I}+\left(\macierz{H}_1\right)^T\macierz{H}_2
\end{displaymath}
are equal to $L_k+p_k \times L_k+p_k$ which is independent of the number of training
of examples.

Similarly to nonregularised variant presented above we can use the
same techniques to solve optimisation problem \eqref{eq:reg-opt-1} and
\eqref{eq:reg-opt-2}. For more details see \citep{fung01proximal}.


\bibliography{my-bibliography}
\end{document}